\documentclass[conference]{IEEEtran}
\IEEEoverridecommandlockouts

\usepackage[utf8]{inputenc} 
\usepackage[T1]{fontenc}    
\usepackage[utf8]{vietnam}
\usepackage[english]{babel}
\usepackage{tipa}
\usepackage{ragged2e}

\usepackage{cite}
\usepackage{hyperref}       
\usepackage{url}            

\usepackage{amsmath,amssymb,amsfonts}
\usepackage{algorithmic}
\usepackage{graphicx}
\usepackage{textcomp}
\usepackage{xcolor}

\usepackage{multirow}
\usepackage{caption}
\usepackage{subcaption}

\usepackage{tabularx}
\usepackage{tkz-tab}

\usepackage{pgfplots}
\usepackage{tikz}
\usepackage{pgfplotstable}
\usepackage{filecontents}

\usepgfplotslibrary{colorbrewer}
\pgfplotsset{compat = 1.15, cycle list/Set1-8} 
\usetikzlibrary{pgfplots.statistics, pgfplots.colorbrewer} 

\definecolor{markercolor}{RGB}{124.9, 255, 160.65}
\pgfplotsset{compat=1.3}
\usetikzlibrary{positioning, fit, calc}  
\tikzset{block/.style={draw, thick, text width=2cm ,minimum height=1.3cm, align=center},   
	line/.style={-latex}     
} 

\tikzset{blocktext/.style={draw, thick, text width=5.2cm ,minimum height=1.3cm, align=center},   
	line/.style={-latex}     
}

\def\BibTeX{{\rm B\kern-.05em{\sc i\kern-.025em b}\kern-.08em
    T\kern-.1667em\lower.7ex\hbox{E}\kern-.125emX}}

\makeatletter
\def\endthebibliography{%
	\def\@noitemerr{\@latex@warning{Empty `thebibliography' environment}}%
	\endlist
}

\newcommand{\linebreakand}{%
\end{@IEEEauthorhalign}
\hfill\mbox{}\par
\mbox{}\hfill\begin{@IEEEauthorhalign}
}
\makeatother

\begin{document}

\title{Can ChatGPT pass the Vietnamese National High School Graduation Examination?}

\author{\IEEEauthorblockN{Xuan-Quy Dao}
\IEEEauthorblockA{
\textit{Eastern International University}\\
quy.dao@eiu.edu.vn}
\and
\IEEEauthorblockN{Ngoc-Bich Le}
\IEEEauthorblockA{
\textit{International University Vietnam}\\
lnbich@hcmiu.edu.vn }
\linebreakand 
\IEEEauthorblockN{Xuan-Dung Phan}
\IEEEauthorblockA{
\textit{Eastern International University}\\
dung.phan@eiu.edu.vn}
\and
\IEEEauthorblockN{Bac-Bien Ngo}
\IEEEauthorblockA{
\textit{Eastern International University}\\
bien.ngo@eiu.edu.vn}
}

\maketitle

\begin{abstract}

This research article highlights the potential of AI-powered chatbots in education and presents the results of using ChatGPT, a large language model, to complete the Vietnamese National High School Graduation Examination (VNHSGE). The study dataset included 30 essays in the literature test case and 1,700 multiple-choice questions designed for other subjects. The results showed that ChatGPT was able to pass the examination with an average score of 6-7, demonstrating the technology's potential to revolutionize the educational landscape. The analysis of ChatGPT performance revealed its proficiency in a range of subjects, including mathematics, English, physics, chemistry, biology, history, geography, civic education, and literature, which suggests its potential to provide effective support for learners. However, further research is needed to assess ChatGPT performance on more complex exam questions and its potential to support learners in different contexts. As technology continues to evolve and improve, we can expect to see the use of AI tools like ChatGPT become increasingly common in educational settings, ultimately enhancing the educational experience for both students and educators.

\end{abstract}


\section{Introduction}

Artificial intelligence (AI) has the potential to revolutionize education. According to Chassignol et al.~\cite{chassignol2018artificial}, AI can transform four areas of education: customized content, innovative teaching strategies, technology-enhanced evaluation, and communication between students and teachers. Zawacki-Richter et al.~\cite{zawacki2019systematic} provided an overview of AI applications in higher education, including profiling and prediction, evaluation and assessment, adaptive systems and personalization, and intelligent tutoring systems. Hwang et al.~\cite{hwang2020vision} suggested potential research topics in AI applications for education. Chen et al.~\cite{chen2020artificial} focused on using AI for administration, instruction, and learning to improve administrative operations, content modification, and learning quality. Dao et al.~\cite{Dao2021} highlighted the potential of generative AI in education to reduce workload and increase learner engagement in online learning. Finally, Nguyen et al.~\cite{Nguyen2021} proposed an online learning platform that incorporates a Vietnamese virtual assistant to assist teachers in presenting lectures to students and to simplify editing without the need for video recording.

Recent advancements in large language models (LLMs) have enabled AI to understand and communicate with humans, creating opportunities for its use in education. LLMs have shown great potential in education, content development, and language translation. The two primary architectures of LLMs are BERT (Bidirectional Encoder Representations from Transformers) and GPT (Generative Pre-trained Transformer). In 2018, Google introduced BERT~\cite{devlin2018bert}, which excelled in various natural language processing (NLP) tasks. OpenAI developed the GPT algorithm~\cite{alec2018improving}, which was trained on extensive unlabeled text datasets. Facebook’s RoBERTa~\cite{liu2019roberta} built on Google’s research, and Google released T5~\cite{raffel2020exploring} in 2019. In 2020, OpenAI created GPT-3~\cite{brown2020language}, which demonstrated exceptional performance in various NLP tasks. Recently, OpenAI developed GPT-4~\cite{OpenAI_gpt_4_report}, a text-to-text machine learning system capable of processing both text and image inputs. GPT-4 has shown human-level performance in many professional and academic criteria, although it may not perform as well as humans in other contexts.

In recent years, there has been an increasing interest in the potential use of chatbots in education. Several studies have explored the potential benefits, concerns, and challenges of integrating chatbots, specifically ChatGPT\footnote{\href{https://chat.openai.com/chat}{https://chat.openai.com/chat}}, into educational settings. Halaweh et al.~\cite{halaweh2023chatgpt} examined the concerns expressed by educators regarding the integration of ChatGPT into educational settings. In another study, Zhai et al.~\cite{Zhai2023} conducted a study to explore the potential impacts of ChatGPT on education by using it to write an academic paper.  Furthermore, Stephen~\cite{atlas2023chatgpt} has discussed potential applications of the ChatGPT language model in higher education, including brainstorming and writing help, professional communication, and individualized learning.  Kasneci et al.~\cite{kasneci2023chatgpt} have discussed the potential benefits and challenges of using large language models in educational settings. 

Studies have examined the ability of ChatGPT, a large language model, to participate in exams. Gerd Kortemeyer~\cite{kortemeyer2023could} evaluated ChatGPT’s performance on a calculus-based physics course and found that it would narrowly pass but exhibited errors of a beginning learner. Katz et al.~\cite{katz2023gpt} evaluated GPT-4’s zero-shot performance on the Uniform Bar Examination (UBE) and found that it outperformed human test-takers and prior models on the multiple-choice component, beating humans in five out of seven subject areas. GPT-4 also scored well on the open-ended essay and performance test components, exceeding the passing threshold for all UBE jurisdictions. Gilson et al.~\cite{gilson2023does} evaluated ChatGPT’s performance on multiple-choice questions from the United States Medical Licensing Examination (USMLE) Step 1 and Step 2 exams and found that its performance decreased as question difficulty increased but provided logical justification for its answer selection. Teo~\cite{Susnjak2022} conducted a study to evaluate ChatGPT’s ability to produce text indistinguishable from human-generated text and exhibit critical thinking skills, suggesting that it has the potential to be used for academic misconduct in online exams. The authors propose using invigilated and oral exams, advanced proctoring techniques, and AI-text output detectors to combat cheating but acknowledge that these solutions may not be foolproof.

The existing literature highlights the potential of chatbots to enhance learning outcomes and provide support to students in various educational settings. However, there is a lack of research investigating their ability to complete high-stakes examinations. Hence, this study aims to bridge this gap in the literature by examining ChatGPT potential to pass the Vietnamese National High School Graduation Examination (VNHSGE). In this paper, we present the results of ChatGPT in completing the VNHSGE exam. The purpose of this study is to answer the question: "Can ChatGPT pass the Vietnamese High School Graduation Examination?" To answer this question, we utilized the VNHSGE dataset~\cite{dao2023vnhsge} which was created from the VNHSGE exam and similar examinations and designed to evaluate the capabilities of LLMs . We evaluated ChatGPT performance on this dataset and analyzed the results to determine the model's accuracy in completing the exam. Furthermore, we will compare the performance of ChatGPT in our test case to other examinations previously conducted by the OpenAI team~\cite{OpenAI_gpt_4_report}. By doing so, this study aims to contribute to a better understanding of the effectiveness of AI-powered chatbots in supporting learners in high-stakes examinations.  

\section{ Methods}

This paper aims is to examine the performance of ChatGPT in completing the VNHSGE exam and provide critical analysis on the results. Through this research, we aim to contribute to the development of AI tools for educational support, and to shed light on the future possibilities of AI in transforming the education landscape.

\subsection{Dataset}

In this study, we utilized the evaluation set of the VNHSGE dataset~\cite{dao2023vnhsge}. The dataset includes questions from the VNHSGE exam for various subjects such as mathematics, English, physics, chemistry, biology, history, geography, civic education, and literature. The VNHSGE dataset was compiled by high school instructors and the Vietnamese Ministry of Education and Training (VMET) and includes both official and illustrative exam questions.

\subsection{Prompt}

In this study, we employed zero-shot learning to evaluate the performance of ChatGPT on the VNHSGE dataset~\textcolor{red}{D}. The dataset consists of pairs of questions~\textcolor{red}{Q} and ground truth solutions~\textcolor{red}{S}. We define the context of words as~\textcolor{red}{P}. The answer~\textcolor{red}{A} of ChatGPT is determined by the following equation
\begin{equation}
	A = f(P, Q)
\end{equation}
where~\textcolor{red}{f} is ChatGPT and takes into account the context~\textcolor{red}{P} and the question~\textcolor{red}{Q}. The context~\textcolor{red}{P} in this case is a specific structure that guides the response of ChatGPT. It instructs ChatGPT to provide the answer in the following format: \{ Choice: “A” or “B” or “C” or “D”;
Explanation: Explain the answer;
The question is: [the actual question] \}. By following this structure, ChatGPT generates its answer~\textcolor{red}{A}, which can be evaluated and compared to the ground truth solution~\textcolor{red}{S}.

\begin{figure*}[ht!]
	\begin{center}
		\begin{tikzpicture}  
			[
			node distance = 1.5cm, 
			]
			\node[block] (a) {Multiple Choice Question};  
			\node[block,right=of a] (b) {New Question};   
			\node[block,right=of b] (c) {ChatGPT OpenAI API};  
			\node[block,right=of c] (d) {Response}; 
			\node[blocktext,align=flush left,font=\small] (f) at ([yshift=2cm]$(c)!0.5!(d)$) {I want you to answer the question in the following structure:  \\ 
				Choice: "A" or "B" or "C" or "D"  \\
				Explanation: Explain the answer \\
				The question is: {}
			};   
			\node[block] (h) at ([yshift=2cm]$(b)!1.0!(b)$) {Context}; 
			\draw[line] (a)-- (b);  
			\draw[line] (b)-- (c) node [midway, above, sloped] (TextNode) {prompt};  
			\draw[line] (c)-- (d);  
			\draw[line] (h)-- (b);  
			\draw[line] (f)-- (h);  
		\end{tikzpicture} 
	\end{center}
	\caption{Prompt to ChatGPT.}
	\label{fig:chatbot_response}
\end{figure*}
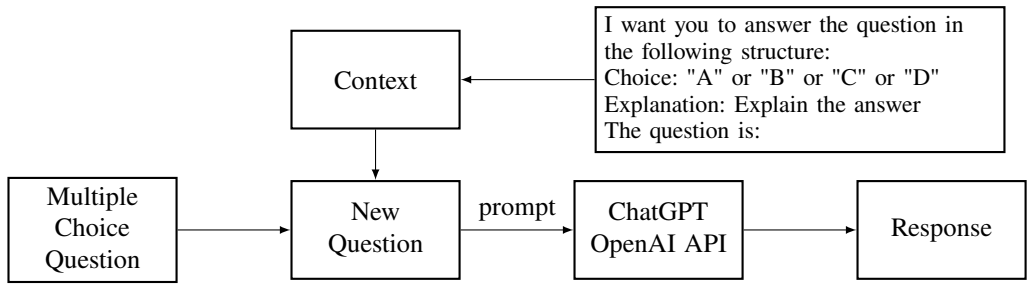

Figure~\ref{fig:chatbot_response} illustrates the process of prompting ChatGPT and retrieving the results. In the case of multiple-choice questions from the VNHSGE dataset, the questions are formatted to align with the expected answer format. The questions are then sent to OpenAI' API.

\subsection{Grading}

To evaluate the performance of ChatGPT in answering questions, we assessed ChatGPT's response by comparing it to the ground truth solution. The evaluation process was conducted using a binary grading system, where ChatGPT's answer was classified as correct or incorrect. The ground truth solution~\textcolor{red}{S} for each question~\textcolor{red}{Q} was determined by a human expert. The answer~\textcolor{red}{A} generated by ChatGPT was then compared to the ground truth solution using the following equation:
\begin{equation}
	G = g(Q, S, A)  
\end{equation}

\section{Results}

\subsection{ChatGPT’s results}

In this study, we aimed to investigate whether ChatGPT, a large language model trained by OpenAI, could pass the VNHSGE exam. To achieve this goal, we compared ChatGPT performance in completing the examination with the correct answer numbers for mathematics, literature, English, physics, chemistry, biology, history, geography, and civic education in the years 2019 to 2023. The detailed results of ChatGPT performance on the examination are provided in\cite{quy2023thpt_a}.  

Table~\ref{tab:correct_answers} presents the correct answer numbers for each subject in the VNHSGE exam from 2019 to 2023. The results show that the number of correct answers varies across subjects and years. The highest number of correct answers in a subject was 43 for English in 2020, while the lowest was 16 for chemistry in 2019. Overall, the number of correct answers in each subject remains relatively stable over the years, with only slight fluctuations.

\begingroup
\renewcommand{\arraystretch}{1.5} 

\begin{table}[ht!]
	\caption{The number of correct answers for multiple-choice subjects}
	\label{tab:correct_answers}
	\centering
	\begin{tabular}{c|c|c|c|c|c|c|c|c|}
		\cline{2-9}
		\multicolumn{1}{l|}{}      & \textbf{Math} & \textbf{Eng} & \textbf{Phy} & \textbf{Che} & \textbf{Bio} & \textbf{His} & \textbf{Geo} & \textbf{Civ} \\ \hline
		\multicolumn{1}{|c|}{2019} & 26                   & 38               & 24               & 16                 & 24               & 17               & 20                 & 24                       \\ \hline
		\multicolumn{1}{|c|}{2020} & 33                   & 43               & 25               & 17                 & 24               & 19               & 21                 & 28                       \\ \hline
		\multicolumn{1}{|c|}{2021} & 30                   & 38               & 24               & 25                 & 21               & 22               & 30                 & 25                       \\ \hline
		\multicolumn{1}{|c|}{2022} & 31                   & 40               & 26               & 19                 & 23               & 24               & 25                 & 33                       \\ \hline
		\multicolumn{1}{|c|}{2023} & 27                   & 39               & 23               & 19                 & 24               & 31               & 27                 & 31                       \\ \hline
	\end{tabular}

\end{table}
\endgroup

\begingroup
\renewcommand{\arraystretch}{1.5} 

\begin{table*}[h!]
	\caption{The scores for subjects on a 10-point scale}
	\label{tab:scores}
	\centering
	\begin{tabular}{c|c|c|c|c|c|c|c|c|c|}
		\cline{2-10}
		\multicolumn{1}{l|}{}         & \textbf{Math} & \textbf{Lit} & \textbf{Eng} & \textbf{Phy} & \textbf{Che} & \textbf{Bio} & \textbf{His} & \textbf{Geo} & \textbf{Civ} \\ \hline
		\multicolumn{1}{|c|}{2019}    & 5.2                  & 7.5                 & 7.6              & 6                & 4                  & 6                & 4.25             & 5                  & 6                        \\ \hline
		\multicolumn{1}{|c|}{2020}    & 6.6                  & 6.89                & 8.6              & 6.25             & 4.25               & 6                & 4.75             & 5.25               & 7                        \\ \hline
		\multicolumn{1}{|c|}{2021}    & 6                    & 7.5                 & 7.6              & 6                & 6.25               & 5.25             & 5.5              & 7.5                & 6.25                     \\ \hline
		\multicolumn{1}{|c|}{2022}    & 6.2                  & 5.63                & 8                & 6.5              & 4.75               & 5.75             & 6                & 6.25               & 8.25                     \\ \hline
		\multicolumn{1}{|c|}{2023}    & 5.4                  & 6.48                & 7.8              & 5.75             & 4.75               & 6                & 7.75             & 6.75               & 7.75                     \\ \hline
		\multicolumn{1}{|c|}{Avg} & 5.88                 & 6.8                 & 7.92             & 6.1              & 4.8                & 5.8              & 5.65             & 6.15               & 7.05                     \\ \hline
	\end{tabular}
\end{table*}
\endgroup

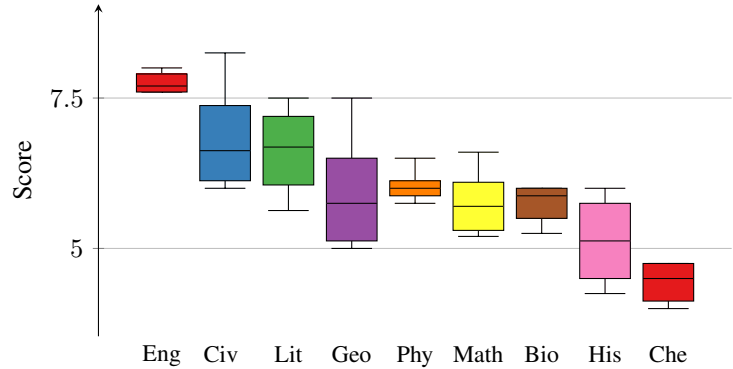
\begin{figure*}[h]
	\begin{center}
		\begin{tikzpicture}
			\pgfplotstableread[col sep=comma]{score}\csvdata
			\pgfplotstabletranspose\datatransposed{\csvdata} 
			\begin{axis}[
				boxplot/draw direction = y,
				x axis line style = {opacity=0},
				axis x line* = bottom,
				height=6cm,
				width=10cm,
				axis y line = left,
				enlarge y limits,
				ymajorgrids,
				xmin=0,
				xmax=10,
				xtick = {1, 2, 3, 4, 5, 6, 7, 8, 9},
				xticklabel style = {align=center, font=\small, rotate=0},
				xticklabels = {
					Eng,			
					Civ ,
					Lit,
					Geo,
					Phy,
					Math,
					Bio,
					His,
					Che
				},
				xtick style = {draw=none}, 
				ylabel = {Score},
				ytick = {0, 2.5, 5.0, 7.5, 10}
				]
				\foreach \n in {1,...,9} {
					\addplot+[boxplot, fill, draw=black] table[y index=\n] {\datatransposed};
				}
			\end{axis}
		\end{tikzpicture}
	\end{center}
	\caption{Average scores and standard deviation of subjects for the years 2019-2023.}
	\label{fig:chatgpt_performance_score}
\end{figure*}

Table~\ref{tab:scores} displays the scores obtained by ChatGPT in the VNHSGE exam across all subjects from 2019 to 2023. The data indicates that the average scores achieved by ChatGPT range from 4.8 to 7.92, with the lowest average score being observed in chemistry, and the highest in English. This highlights ChatGPT diverse performance across the range of subjects, with some subjects yielding higher scores than others. Figure~\ref{fig:chatgpt_performance_score} presents the average results of the subjects for each year from 2019 to 2023, along with the corresponding standard deviation bars. The average results were computed based on the data presented in Table~\ref{tab:scores}, which shows ChatGPT scores for mathematics, English, physics, chemistry, biology, history, geography, civic education, and literature for each year. English consistently had the highest average score, ranging from 7.6 in 2019 to 8.6 in 2020, followed by civic education with scores ranging from 6.0 in 2019 to 8.25 in 2022. In contrast, chemistry had the lowest average score, ranging from 4.0 in 2019 to 4.75 in 2022. Biology and history also had relatively low average scores, ranging from 5.25 to 6.0. The results reveal that the performance of ChatGPT in chemistry has remained relatively stable across all four years, with the exception of 2021. Conversely, there is a noticeable variability in ChatGPT scores for subjects such as history, civic education, and geography, suggesting that its performance is less consistent in these areas. The data analysis shows a low degree of variability in ChatGPT scores in subjects like English, literature, physics, and biology, indicating a relatively consistent performance in these subjects. It is worth noting that the variability in ChatGPT scores could be attributed to the complexity of questions and topics covered in these subjects, which may require a deeper understanding of cultural and historical contexts. Overall, the results suggest that ChatGPT performance in certain subjects might require further development to achieve a more consistent and stable performance level, which could be vital for passing the VNHSGE exam.

\subsection{Comparison of ChatGPT's performance in VNHSGE and other exams}

\begin{figure*}[h!]
	\begin{center}
		\begin{tikzpicture}
			\begin{axis}[
				ylabel={\pgfmathprintnumber{\tick}\%},
				legend style={at={(0.5,1.125)}, 	
					anchor=north,legend columns=-1},
				symbolic x coords={
					AP Calculus BC,
					AMC 12,
					Codeforces Rating,
					AP English Literature,
					AMC 10,
					Uniform Bar Exam,
					AP English Language,
					AP Chemistry,
					GRE Quantitative,
					AP Physics 2,
					USABO Semifinal 2020,
					AP Macroeconomics,
					LSAT,
					AP Statistics,
					VNHSGE Chemistry,
					GRE Writing,
					VNHSGE History,
					VNHSGE Biology,
					VNHSGE Mathematics,
					AP Microeconomics,
					AP Biology,
					VNHSGE Physics,
					VNHSGE Geography,
					GRE Verbal,
					AP World History,
					VNHSGE Literature,
					SAT Math,
					VNHSGE Civic Education,
					AP US History,
					AP US Government,
					VNHSGE English,
					AP Psychology,
					AP Art History,
					SAT EBRW,
					AP Environmental Science,
				},
				xtick=data,
				hide axis,
				x tick label style={rotate=60,anchor=east},
				ybar,
				bar width=7.5pt,
				ymin=0,
				ymax=100,
				width=\textwidth, 
				enlarge x limits={abs=0.5*\pgfplotbarwidth},					
				height=9cm, width=16cm,
				]
				
				\addplot [fill=blue] coordinates {
					(AP Microeconomics,0)
				};	
					
				\addplot [fill=yellow] coordinates {
					(AP Environmental Science,0)	
				};		
				\legend{GPT-3.5, ChatGPT}				
			\end{axis}
			\begin{axis}[
				ylabel={},
				symbolic x coords={
					AP Calculus BC,
					AMC 12,
					Codeforces Rating,
					AP English Literature,
					AMC 10,
					Uniform Bar Exam,
					AP English Language,
					AP Chemistry,
					GRE Quantitative,
					AP Physics 2,
					USABO Semifinal 2020,
					AP Macroeconomics,
					LSAT,
					AP Statistics,
					VNHSGE Chemistry,
					GRE Writing,
					VNHSGE History,
					VNHSGE Biology,
					VNHSGE Mathematics,
					AP Microeconomics,
					AP Biology,
					VNHSGE Physics,
					VNHSGE Geography,
					GRE Verbal,
					AP World History,
					VNHSGE Literature,
					SAT Math,
					VNHSGE Civic Education,
					AP US History,
					AP US Government,
					VNHSGE English,
					AP Psychology,
					AP Art History,
					SAT EBRW,
					AP Environmental Science,
				},
				xtick=data,
				x tick label style={rotate=60,anchor=east},
				yticklabel={\pgfmathprintnumber{\tick}\%},
				ybar,
				bar width=7.5pt,
				ymin=0,
				ymax=100,
				width=\textwidth, 
				enlarge x limits={abs=0.5*\pgfplotbarwidth},					
				height=9cm, width=16cm,
				]
				\addplot [fill=blue] coordinates {
					(AP Calculus BC,1)
					(AMC 12,4)
					(Codeforces Rating,5)
					(AP English Literature,8)
					(AMC 10,10)
					(Uniform Bar Exam,10)
					(AP English Language,14)
					(AP Chemistry,22)
					(GRE Quantitative,25)
					(AP Physics 2,30)
					(USABO Semifinal 2020,31)
					(AP Macroeconomics,33)
					(LSAT,40)
					(AP Statistics,40)
					(VNHSGE Chemistry,0)
					(GRE Writing,54)
					(VNHSGE History,0)
					(VNHSGE Biology,0)
					(VNHSGE Mathematics,0)
					(AP Microeconomics,60)
					(AP Biology,62)
					(VNHSGE Physics,0)
					(VNHSGE Geography,0)
					(GRE Verbal,63)
					(AP World History,65)
					(VNHSGE Literature,0)
					(SAT Math,70)
					(VNHSGE Civic Education,0)
					(AP US History,74)
					(AP US Government,77)
					(VNHSGE English,0)
					(AP Psychology,85)
					(AP Art History,86)
					(SAT EBRW,86)
					(AP Environmental Science,90)	
				};					
			\end{axis}
			\begin{axis}[
				ylabel={Percentage},
				symbolic x coords={
					AP Calculus BC,
					AMC 12,
					Codeforces Rating,
					AP English Literature,
					AMC 10,
					Uniform Bar Exam,
					AP English Language,
					AP Chemistry,
					GRE Quantitative,
					AP Physics 2,
					USABO Semifinal 2020,
					AP Macroeconomics,
					LSAT,
					AP Statistics,
					VNHSGE Chemistry,
					GRE Writing,
					VNHSGE History,
					VNHSGE Biology,
					VNHSGE Mathematics,
					AP Microeconomics,
					AP Biology,
					VNHSGE Physics,
					VNHSGE Geography,
					GRE Verbal,
					AP World History,
					VNHSGE Literature,
					SAT Math,
					VNHSGE Civic Education,
					AP US History,
					AP US Government,
					VNHSGE English,
					AP Psychology,
					AP Art History,
					SAT EBRW,
					AP Environmental Science,
				},
				x tick label style={rotate=60,anchor=east},
				hide axis,
				ybar,
				bar width=7.5pt,
				ymin=0,
				ymax=100,
				width=\textwidth, 
				enlarge x limits={abs=0.5*\pgfplotbarwidth},					
				height=9cm, width=16cm,
				]	
				\addplot [fill=yellow] coordinates {
					(AP Calculus BC,0)
					(AMC 12,0)
					(Codeforces Rating,0)
					(AP English Literature,0)
					(AMC 10,0)
					(Uniform Bar Exam,0)
					(AP English Language,0)
					(AP Chemistry,0)
					(GRE Quantitative,0)
					(AP Physics 2,0)
					(USABO Semifinal 2020,0)
					(AP Macroeconomics,0)
					(LSAT,0)
					(AP Statistics,00)
					(VNHSGE Chemistry,48)
					(GRE Writing,0)
					(VNHSGE History,56.5)
					(VNHSGE Biology,58)
					(VNHSGE Mathematics,58.8)
					(AP Microeconomics,0)
					(AP Biology,0)
					(VNHSGE Physics,61)
					(VNHSGE Geography,61.5)
					(GRE Verbal,0)
					(AP World History,0)
					(VNHSGE Literature,67.96)
					(SAT Math,0)
					(VNHSGE Civic Education,70.5)
					(AP US History,0)
					(AP US Government,0)
					(VNHSGE English,79)
					(AP Psychology,0)
					(AP Art History,0)
					(SAT EBRW,0)
					(AP Environmental Science,0)	
				};
			\end{axis}
		\end{tikzpicture}
	\end{center}
	\caption{ChatGPT performance in Vietnamese examinations and other examinations.}
	\label{fig:chatgpt_exam}
\end{figure*}
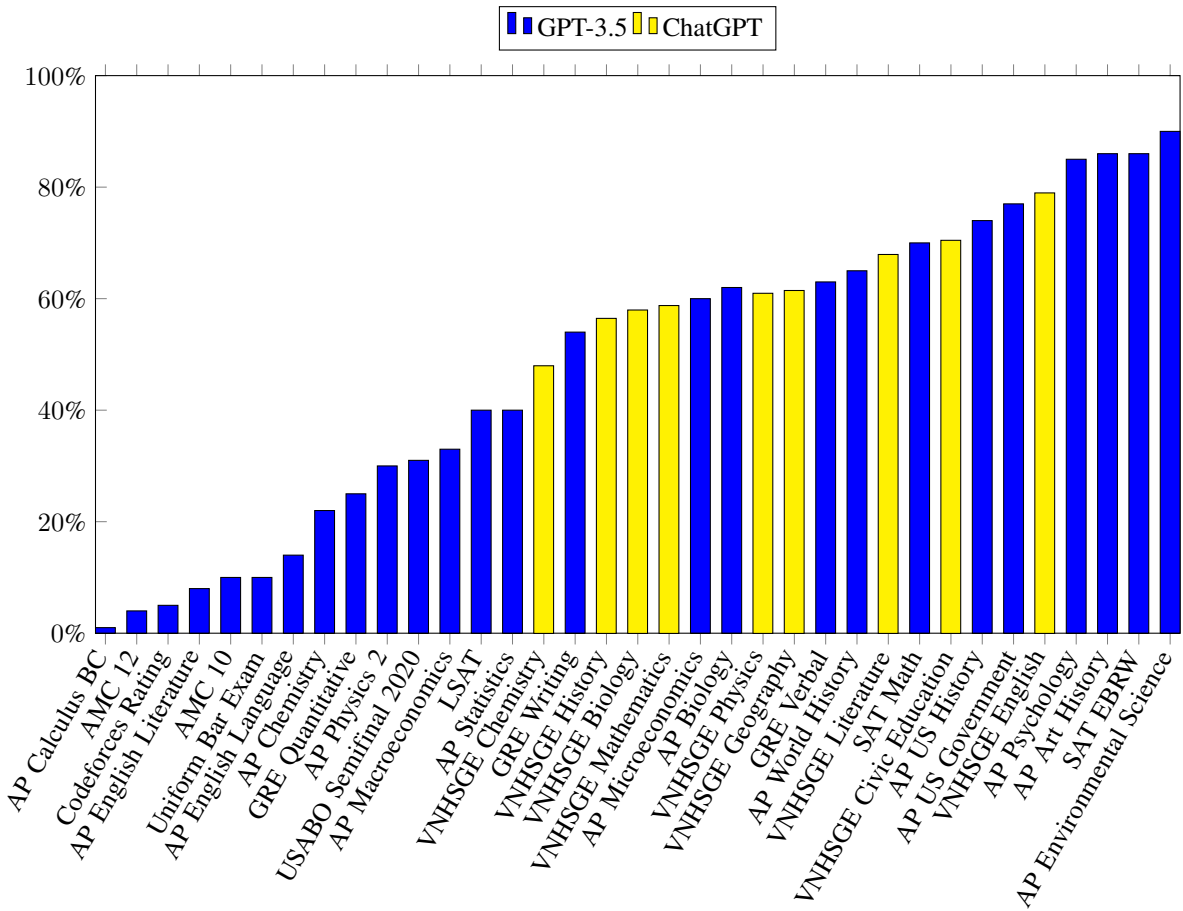

Figure~\ref{fig:chatgpt_exam} depicts ChatGPT performance in various test cases, including the VNHSGE exam and other examinations~\cite{OpenAI_gpt_4_report}. Overall, the results show that ChatGPT performance in the VNHSGE exam is comparable to its performance in other examinations, such as SAT Math, AP Biology, and AP Physics. However, there are some differences in its performance across various test cases. For example, in the mathematics test case, ChatGPT achieved a lower score on the VNHSGE mathematics test compared to the SAT Math. This difference could be due to the higher level of analytical and critical thinking skills required in the VNHSGE mathematics test, which is also conducted in Vietnamese.

In contrast, ChatGPT performance on the literature test case was higher than that of the GRE Writing examination. This success can be attributed to ChatGPT ability to search and summarize content effectively, as well as its training on a vast corpus of paragraphs, enabling it to produce a well-structured essay. Similarly, ChatGPT achieved a significantly higher score on the VNHSGE English test than on the AP English language examination. This difference could be due to the level of difficulty of the tests, with the VNHSGE English test being comparatively easier than the AP language examination.

In the physics test case, ChatGPT accuracy in the VNHSGE physics exam was considerably higher than that in the AP Physics 2 exam. In the chemistry test case, ChatGPT achieved its lowest score in the VNHSGE chemistry examination, but its performance was still significantly better than that in the AP Chemistry examination. The difference in performance could be due to the varying content and format of the exams, the level of difficulty of the assessment questions, and ChatGPT level of exposure to each exam's content.

Regarding the biology test case, ChatGPT performance across the VNHSGE biology exam and AP Biology examinations was relatively similar, achieving comparable accuracy scores of approximately 60\%. Similarly, there were no significant differences observed between ChatGPT performance on the VNHSGE history examination and the AP World History examination, with both tests producing scores of approximately 56.5\% and 61\%, respectively.

Unfortunately, there is a lack of data from the two subjects, geography and civic education, which makes it impossible to draw any conclusions about ChatGPT performance in these areas. In summary, the results suggest that ChatGPT performance varies across different test cases, and it is important to consider factors such as test difficulty, content, and language when evaluating its performance.

\subsection{ChatGPT and Vietnamese Student Score Spectrums}

\begingroup
\renewcommand{\arraystretch}{1.5} 
\begin{table*}[ht!]
	\caption{Average score and Most reached score of Vietnamese students}
	\label{tabl:AMVS}
	\resizebox{\textwidth}{!}{%
		\begin{tabular}{c|cc|cc|cc|cc|cc|cc|cc|cc|cc|}
			\cline{2-19}
			\multicolumn{1}{l|}{}               & \multicolumn{2}{c|}{\textbf{Math}} & \multicolumn{2}{c|}{\textbf{Lit}} & \multicolumn{2}{c|}{\textbf{Eng}} & \multicolumn{2}{c|}{\textbf{Phy}} & \multicolumn{2}{c|}{\textbf{Che}} & \multicolumn{2}{c|}{\textbf{Bio}} & \multicolumn{2}{c|}{\textbf{His}} & \multicolumn{2}{c|}{\textbf{Geo}} & \multicolumn{2}{c|}{\textbf{Civ}} \\ \hline
			\multicolumn{1}{|c|}{\textbf{2019}} & \multicolumn{1}{c|}{5.64}      & 6.4      & \multicolumn{1}{c|}{5.49}       & 6      & \multicolumn{1}{c|}{4.36}    & 3.2    & \multicolumn{1}{c|}{5.57}    & 6.25   & \multicolumn{1}{c|}{5.35}     & 6       & \multicolumn{1}{c|}{4.68}    & 4.5    & \multicolumn{1}{c|}{4.3}     & 3.75   & \multicolumn{1}{c|}{6}        & 6       & \multicolumn{1}{c|}{7.37}        & 7.75       \\ \hline
			\multicolumn{1}{|c|}{\textbf{2020}} & \multicolumn{1}{c|}{6.67}      & 7.8      & \multicolumn{1}{c|}{6.61}       & 7      & \multicolumn{1}{c|}{4.58}    & 3.4    & \multicolumn{1}{c|}{6.72}    & 7.75   & \multicolumn{1}{c|}{6.71}     & 7.75    & \multicolumn{1}{c|}{5.6}     & 5.25   & \multicolumn{1}{c|}{5.19}    & 4.5    & \multicolumn{1}{c|}{6.78}     & 7.25    & \multicolumn{1}{c|}{8.14}        & 8.75       \\ \hline
			\multicolumn{1}{|c|}{\textbf{2021}} & \multicolumn{1}{c|}{6.61}      & 7.8      & \multicolumn{1}{c|}{6.47}       & 7      & \multicolumn{1}{c|}{5.84}    & 4      & \multicolumn{1}{c|}{6.56}    & 7.5    & \multicolumn{1}{c|}{6.63}     & 7.75    & \multicolumn{1}{c|}{5.51}    & 5.25   & \multicolumn{1}{c|}{4.97}    & 4      & \multicolumn{1}{c|}{6.96}     & 7       & \multicolumn{1}{c|}{8.37}        & 9.25       \\ \hline
			\multicolumn{1}{|c|}{\textbf{2022}} & \multicolumn{1}{c|}{6.47}      & 7.8      & \multicolumn{1}{c|}{6.51}       & 7      & \multicolumn{1}{c|}{5.15}    & 3.8    & \multicolumn{1}{c|}{6.72}    & 7.25   & \multicolumn{1}{c|}{6.7}      & 8       & \multicolumn{1}{c|}{5.02}    & 4.5    & \multicolumn{1}{c|}{6.34}    & 7      & \multicolumn{1}{c|}{6.68}     & 7       & \multicolumn{1}{c|}{8.03}        & 8.5        \\ \hline
		\end{tabular}
	}
\end{table*}
\endgroup

In this section, ChatGPT results are compared with those of Vietnamese students in the years 
\href{https://moet.gov.vn/tintuc/Pages/tin-tong-hop.aspx?ItemID=6111}{2019},~
\href{https://moet.gov.vn/tintuc/Pages/tin-tong-hop.aspx?ItemID=6879}{2020},~
\href{https://moet.gov.vn/tintuc/Pages/tin-tong-hop.aspx?ItemID=7451}{2021},~
\href{https://vietnamnet.vn/pho-diem-cac-mon-thi-tot-nghiep-thpt-2022-2042421.html}{2022}. The score distribution serves as an indicator for evaluating the performance of candidates in exams. Each year, Vietnam Ministry of Education and Training releases a chart depicting the score distribution for each subject. This distribution helps assess the candidates' competence and gauge the difficulty level of the exams, thus providing an evaluation of the applicants' proficiency. We have gathered the score distributions from 2019 to 2022, allowing us to compare the performance of ChatGPT with that of Vietnamese students. To facilitate this comparison, Table~\ref{tabl:AMVS} presents the average score (AVS) and the most reached score (MVS) by Vietnamese students. For example, in 2019, the AVS and MVS for mathematics are 5.64 and 6.4, respectively.

\subsubsection{Mathematics}
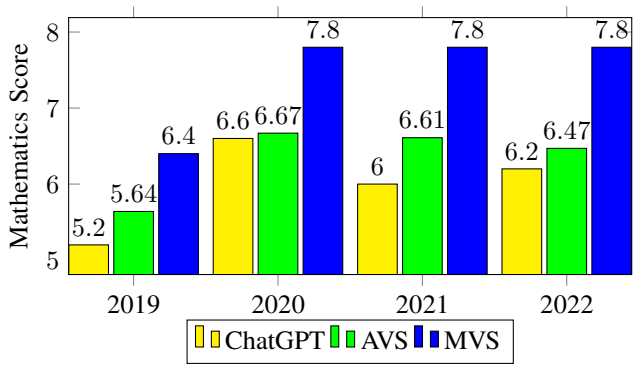
\begin{figure}[ht!]
	\begin{center}
		\begin{tikzpicture}  
			\begin{axis}  
				[  
				ybar, 
				bar width=15pt,
				enlargelimits=0.15, 
				legend style={at={(0.5,-0.175)}, 	
					anchor=north,legend columns=-1},       
				ylabel={Mathematics Score}, 
				symbolic x coords={
					2019,
					2020,
					2021,
					2022,
				},  
				xtick=data,  
				nodes near coords,  
				nodes near coords align={vertical},  
				width=0.5\textwidth, 
				height=5cm, 
				]  
				\addplot [fill=yellow] coordinates {
					(2019,5.2)
					(2020,6.6)
					(2021,6)
					(2022,6.2)
				};
				\addplot [fill=green] coordinates {
					(2019,5.64)
					(2020,6.67)
					(2021,6.61)
					(2022,6.47)	
				};
				\addplot [fill=blue] coordinates {
					(2019,6.4)
					(2020,7.8)
					(2021,7.8)
					(2022,7.8)				
				};
				\legend{ChatGPT, AVS, MVS}
			\end{axis}  
		\end{tikzpicture}
	\caption{Comparison in mathematics score.}
	\label{fig:score_math}
	\end{center}
\end{figure}

In this section, we investigated the performance of ChatGPT in mathematics in comparison to Vietnamese students. The data collected from 2019 to 2022 were analyzed and presented in Figure~\ref{fig:score_math}. The results indicate that the Mathematics scores of ChatGPT ranged from 5.2 to 6.6, which is lower than the majority of Vietnamese students. These findings suggest that while ChatGPT has shown impressive capabilities in natural language processing, its performance in mathematics is still suboptimal. Further investigation is warranted to understand the factors contributing to ChatGPT performance in mathematics and to improve its abilities in this area. 

\subsubsection{Literature}

\begin{figure}[ht!]
	\begin{center}
		\begin{tikzpicture}  
			\begin{axis}  
				[  
				ybar, 
				bar width=15pt,
				enlargelimits=0.15, 
				legend style={at={(0.5,-0.175)}, 	
					anchor=north,legend columns=-1},       
				ylabel={Literature Score}, 
				symbolic x coords={
					2019,
					2020,
					2021,
					2022,
				},  
				xtick=data,  
				nodes near coords,  
				nodes near coords align={vertical},  
				width=0.5\textwidth, 
				height=5cm, 
				]  
				\addplot [fill=yellow] coordinates {
					(2019,7.5)
					(2020,6.89)
					(2021,7.5)
					(2022,5.63)
				};
				\addplot [fill=green] coordinates {
					(2019,5.49)
					(2020,6.61)
					(2021,6.47)
					(2022,6.51)		
				};
				\addplot [fill=blue] coordinates {
					(2019,6)
					(2020,7)
					(2021,7)
					(2022,7)
				};
				\legend{ChatGPT, AVS, MVS} 
			\end{axis}  
		\end{tikzpicture}
	\end{center}
	\caption{Comparison in literature.}
	\label{fig:core_literature}
\end{figure}
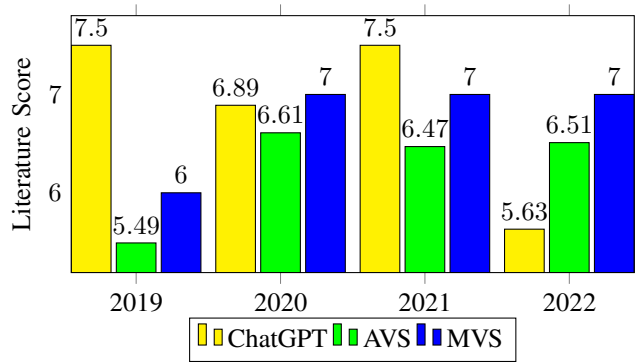

This section examined the performance of ChatGPT in the literature component of Vietnam's national high school graduation exam, illustrated in Figure~\ref{fig:core_literature}. The literature scores of ChatGPT ranged from 5.63 to 7.5, with an average score of 6.68. A comparison between ChatGPT performance and Vietnamese students in the literature test was conducted, and the results were intriguing. ChatGPT scores were consistently higher than the majority of Vietnamese students, except in 2022, where the results were slightly lower. This finding highlights the potential of language models to excel in tasks that require a deep understanding of textual content. Literature is a crucial component of the national high school graduation exam in Vietnam, and the performance of ChatGPT in this subject is critical in determining its ability to pass the exam. However, it is important to note that ChatGPT performance may vary depending on factors such as test difficulty, content, and language. Further research is needed to explore these factors and assess the reliability of ChatGPT performance in literature and other subjects.

\subsubsection{English}

\begin{figure}[ht!]
	\begin{center}
		\begin{tikzpicture}  
			\begin{axis}  
				[  
				ybar, 
				bar width=15pt,
				enlargelimits=0.15, 
				legend style={at={(0.5,-0.175)}, 	
					anchor=north,legend columns=-1},       
				ylabel={English Score}, 
				symbolic x coords={
					2019,
					2020,
					2021,
					2022,
				},  
				xtick=data,  
				nodes near coords,  
				nodes near coords align={vertical},  
				width=0.5\textwidth, 
				height=5cm, 
				]  
				\addplot [fill=yellow] coordinates {
					(2019,7.6)
					(2020,8.6)
					(2021,7.6)
					(2022,8)
				};
				\addplot [fill=green] coordinates {
					(2019,4.36)
					(2020,4.58)
					(2021,5.84)
					(2022,5.15)						
				};
				\addplot [fill=blue] coordinates {
					(2019,3.2)
					(2020,3.4)
					(2021,4)
					(2022,3.8)
				};
				\legend{ChatGPT, AVS, MVS} 
			\end{axis}  
		\end{tikzpicture}  
	\end{center}
	\caption{Comparison in English.}
	\label{fig:core_english}
\end{figure}
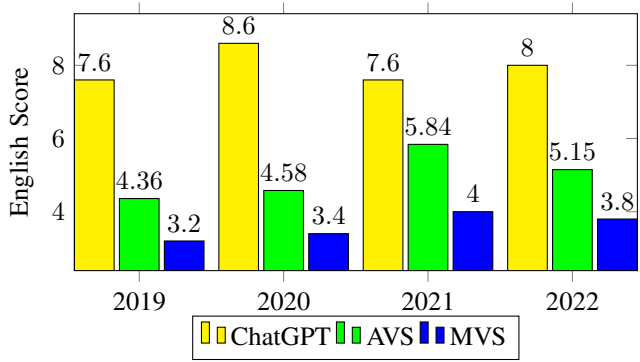

In this section, we examined the English language proficiency of ChatGPT in comparison to Vietnamese students. The English scores of Vietnamese students were compared to those of ChatGPT in the period from 2019 to 2022, and the results were presented in Figure~\ref{fig:core_english}. The findings indicate that ChatGPT achieved a score for the English subject that ranged from 7.8 to 8.6. Notably, the results showed that ChatGPT performed better than the majority of Vietnamese students in English language proficiency. 

It is worth noting that ChatGPT superior performance in English proficiency compared to Vietnamese students could be attributed to several factors, such as its access to a vast corpus of English texts and resources, as well as its ability to learn and adapt from a wide range of language use cases. However, it is important to recognize that the comparison between ChatGPT and Vietnamese students is not entirely straightforward, as ChatGPT is a language model and does not possess the same cognitive and cultural background as human learners. Moreover, the study did not take into account other variables that might influence language learning, such as socio-economic status, age, and motivation, which could affect the English proficiency of Vietnamese students.

Overall, the results of this study demonstrate the potential of large language models like ChatGPT to achieve high levels of proficiency in foreign languages such as English. The study highlights the need for further research to explore the use of such models in language learning and teaching contexts and the implications of their use for learners and educators.

\subsubsection{Physics}

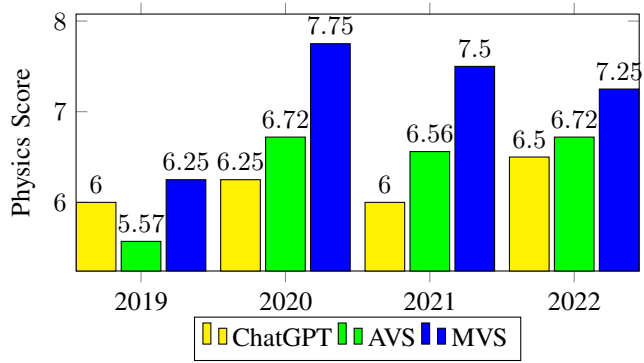
\begin{figure}[ht!]
	\begin{center}
		\begin{tikzpicture}  
			\begin{axis}  
				[  
				ybar, 
				bar width=15pt,
				enlargelimits=0.15, 
				legend style={at={(0.5,-0.1775)}, 	
					anchor=north,legend columns=-1},          
				ylabel={Physics Score}, 
				symbolic x coords={
					2019,
					2020,
					2021,
					2022,
				},  
				xtick=data,  
				nodes near coords,  
				nodes near coords align={vertical},  
				width=0.5\textwidth, 
				height=5cm, 
				]  
				
				\addplot [fill=yellow] coordinates {
					(2019,6.0)
					(2020,6.25)
					(2021,6.0)
					(2022,6.5)
				};
				\addplot [fill=green] coordinates {
					(2019,5.57)
					(2020,6.72)
					(2021,6.56)
					(2022,6.72)	
				};
				\addplot [fill=blue] coordinates {
					(2019,6.25)
					(2020,7.75)
					(2021,7.5)
					(2022,7.25)	
				};
				\legend{ChatGPT, AVS, MVS} 
			\end{axis}  
		\end{tikzpicture} 
	\end{center}
	\caption{Comparison in physics.}
	\label{fig:natural_phy}
\end{figure}

This section investigates the performance of ChatGPT in comparison to Vietnamese students in physics. The results are presented in Figure~\ref{fig:natural_phy}, which displays the physics scores of Vietnamese students compared to ChatGPT over the period of 2019-2022. The findings reveal that ChatGPT score for physics subject varied from 5.75 to 6.5. Interestingly, the results also indicate that ChatGPT mostly achieved lower scores compared to the majority of Vietnamese students. This is consistent with previous research that has shown that language models like ChatGPT may not perform as well as humans on tasks that require reasoning or understanding of physical concepts. Further investigation is warranted to identify the factors contributing to the observed differences in performance between ChatGPT and Vietnamese students.

\subsubsection{Chemistry}

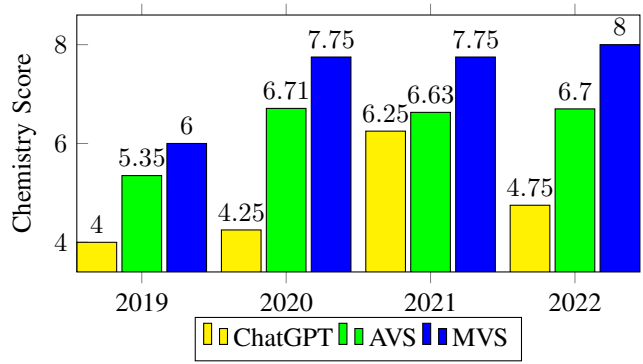
\begin{figure}[ht!]
	\begin{center} 
		\begin{tikzpicture}  
			\begin{axis}  
				[  
				ybar, 
				bar width=15pt,
				enlargelimits=0.15, 
				legend style={at={(0.5,-0.175)}, 	
					anchor=north,legend columns=-1},       
				ylabel={Chemistry Score}, 
				symbolic x coords={
					2019,
					2020,
					2021,
					2022,
				},  
				xtick=data,  
				nodes near coords,  
				nodes near coords align={vertical},  
				width=0.5\textwidth, 
				height=5cm, 
				]  
				\addplot [fill=yellow] coordinates {
					(2019,4)
					(2020,4.25)
					(2021,6.25)
					(2022,4.75)
				};
				\addplot [fill=green] coordinates {
					(2019,5.35)
					(2020,6.71)
					(2021,6.63)
					(2022,6.7)
				};
				\addplot [fill=blue] coordinates {
					(2019,6)
					(2020,7.75)
					(2021,7.75)
					(2022,8)					
				};
				\legend{ChatGPT, AVS, MVS} 
			\end{axis}  
		\end{tikzpicture} 
	\end{center}
	\caption{Comparison in chemistry.}
	\label{fig:natural_che}
\end{figure}

Figure~\ref{fig:natural_che} presents the results of the chemistry subject scores of Vietnamese students compared to ChatGPT in the years 2019 to 2022. The findings reveal that ChatGPT score for the chemistry subject varied between 4.0 to 6.25, indicating a range of performance across the years. Interestingly, ChatGPT performance is consistently lower than the majority of Vietnamese students in chemistry. The results suggest that although ChatGPT has achieved competitive scores in other subjects, its performance in chemistry falls below the expectations of the average Vietnamese student. Further research could investigate the possible factors that contribute to this discrepancy and explore ways to improve ChatGPT performance in chemistry.

\subsubsection{Biology}

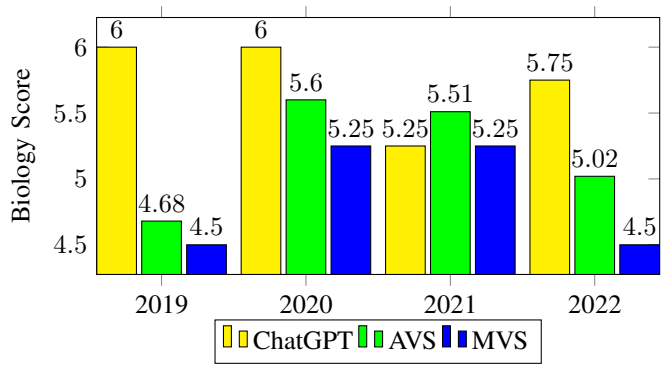
\begin{figure}[ht!]
	\begin{center}
		\begin{tikzpicture}  
			\begin{axis}  
				[  
				ybar, 
				bar width=15pt,
				enlargelimits=0.15, 
				legend style={at={(0.5,-0.175)}, 	
					anchor=north,legend columns=-1},       
				ylabel={Biology Score}, 
				symbolic x coords={
					2019,
					2020,
					2021,
					2022,
				},  
				xtick=data,  
				nodes near coords,  
				nodes near coords align={vertical},  
				width=0.5\textwidth, 
				height=5cm, 
				]  
				\addplot [fill=yellow] coordinates {
					(2019,6)
					(2020,6)
					(2021,5.25)
					(2022,5.75)
				};
				\addplot [fill=green] coordinates {
					(2019,4.68)
					(2020,5.6)
					(2021,5.51)
					(2022,5.02)				
				};
				\addplot [fill=blue] coordinates {
					(2019,4.5)
					(2020,5.25)
					(2021,5.25)
					(2022,4.5)
				};
				\legend{ChatGPT, AVS, MVS} 
			\end{axis}  
		\end{tikzpicture} 
	\end{center}
	\caption{Comparison in biology.}
	\label{fig:natural bio}
\end{figure}

This section assesses the performance of ChatGPT in the biology subject compared to Vietnamese students. Figure~\ref{fig:natural bio} displays the biology subject scores of Vietnamese students compared to ChatGPT from 2019 to 2022. The results indicate that ChatGPT scores for biology ranged from 5.25 to 6.00, with an average score of 5.63. Interestingly, ChatGPT was able to achieve higher results than the majority of Vietnamese students, with only a small proportion of students outscoring the language model. These findings suggest that ChatGPT has the potential to perform well in biology subjects and may serve as a useful tool for educational purposes. However, further research is needed to investigate the factors that contribute to the success of ChatGPT in biology and how it can be integrated into the education system to support student learning.

\subsubsection{History}

\begin{figure}[ht!]
	\begin{center}
		\begin{tikzpicture}  
			\begin{axis}  
				[  
				ybar, 
				bar width=15pt,
				enlargelimits=0.15, 
				legend style={at={(0.5,-0.175)}, 	
					anchor=north,legend columns=-1},       
				ylabel={History Score}, 
				symbolic x coords={
					2019,
					2020,
					2021,
					2022,
				},  
				xtick=data,  
				nodes near coords,  
				nodes near coords align={vertical},  
				width=0.5\textwidth, 
				height=5cm, 
				]  
				\addplot [fill=yellow] coordinates {
					(2019,4.25)
					(2020,4.75)
					(2021,5.5)
					(2022,6)
				};
				\addplot [fill=green] coordinates {
					(2019,4.3)
					(2020,5.19)
					(2021,4.97)
					(2022,6.34)	
				};
				\addplot [fill=blue] coordinates {
					(2019,3.75)
					(2020,4.5)
					(2021,4)
					(2022,7)
				};
				\legend{ChatGPT, AVS, MVS} 
			\end{axis}  
		\end{tikzpicture}
	\end{center}
	\caption{Comparison in history.}
	\label{fig:social_history}
\end{figure}
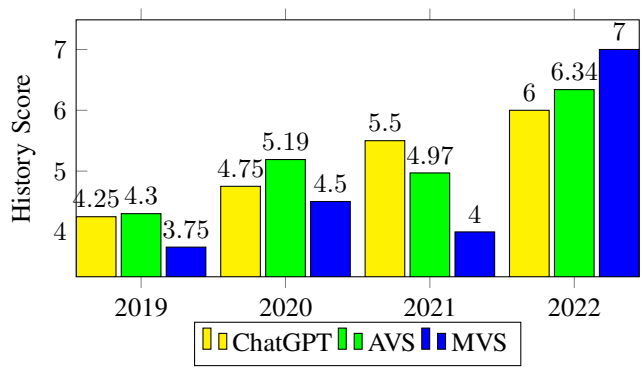

Figure~\ref{fig:social_history} presents a comparison of the history subject scores of Vietnamese students to those of ChatGPT from 2019 to 2022. The results reveal that the ChatGPT score for the history subject ranged from 4.25 to 7.75, indicating a significant variation across the years. Notably, ChatGPT achieved higher results than the majority of Vietnamese students in most cases. These findings suggest that ChatGPT proficiency in history is commendable and may reflect the potential of language models to perform well in various subjects. However, further investigation is needed to identify the factors that contribute to ChatGPT success and to ensure that its performance aligns with the educational standards and requirements of the Vietnamese curriculum.

\subsubsection{Geography}

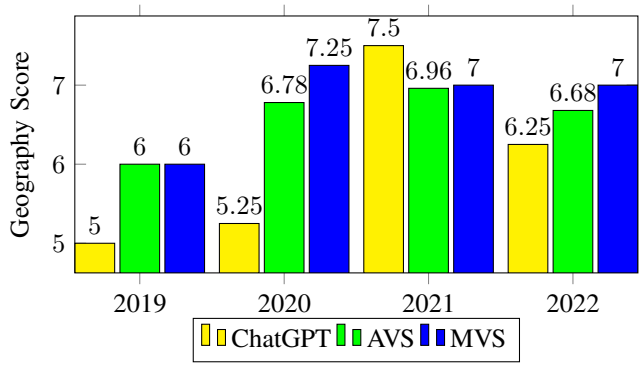
\begin{figure}[ht!]
	\begin{center}
		\begin{tikzpicture}  
			\begin{axis}  
				[  
				ybar, 
				bar width=15pt,
				enlargelimits=0.15, 
				legend style={at={(0.5,-0.175)}, 	
					anchor=north,legend columns=-1},       
				ylabel={Geography Score}, 
				symbolic x coords={
					2019,
					2020,
					2021,
					2022,
				},  
				xtick=data,  
				nodes near coords,  
				nodes near coords align={vertical},  
				width=0.5\textwidth, 
				height=5cm, 
				]  
				\addplot [fill=yellow] coordinates {
					(2019,5)
					(2020,5.25)
					(2021,7.5)
					(2022,6.25)
				};
				\addplot [fill=green] coordinates {
					(2019,6)
					(2020,6.78)
					(2021,6.96)
					(2022,6.68)			
				};
				\addplot [fill=blue] coordinates {
					(2019,6)
					(2020,7.25)
					(2021,7)
					(2022,7)		
				};
				\legend{ChatGPT, AVS, MVS} 
			\end{axis}  
		\end{tikzpicture} 
	\end{center}
	\caption{Comparison in geography.}
	\label{fig:social_geography}
\end{figure}

This section presents a comparison between the performance of ChatGPT and Vietnamese students in the geography subject from 2019 to 2022, as displayed in Figure~\ref{fig:social_geography}. The results reveal that ChatGPT scores in geography ranged from 5.0 to 7.5. Interestingly, in 2021, ChatGPT score surpassed that of most Vietnamese students, yet, in general, ChatGPT achieved lower results than the majority of Vietnamese students. One possible explanation for ChatGPT lower performance is its inability to analyze images and charts. As ChatGPT is based on GPT-3.5, it lacks the capability to read and analyze images or charts, rendering it unable to answer questions related to the geographical Atlas of Vietnam, which accounts for approximately 50\% of questions in the geography exam. To address this issue, we plan to evaluate the performance of GPT-4 to examine its improvement. Moreover, further investigation is necessary to identify the specific factors contributing to the differences in performance between ChatGPT and Vietnamese students in geography. Overall, our findings indicate that while ChatGPT has shown potential in some subjects, it still requires improvement to match the performance of human learners in certain areas. The results of this study highlight the need for future research to enhance the abilities of language models in analyzing and interpreting visual information.

\subsubsection{Civic Education}

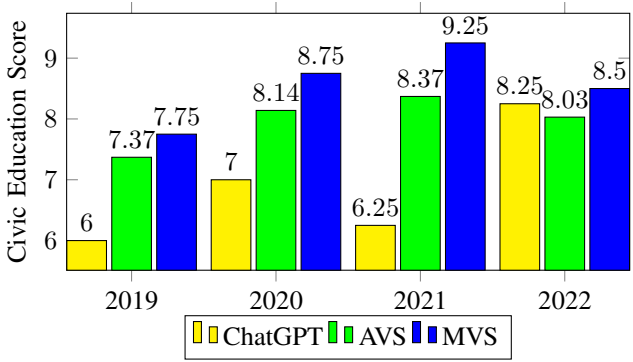
\begin{figure}[ht!]
	\begin{center}
		\begin{tikzpicture}  
			\begin{axis}  
				[  
				ybar, 
				bar width=15pt,
				enlargelimits=0.15, 
				legend style={at={(0.5,-0.175)}, 	
					anchor=north,legend columns=-1},       
				ylabel={Civic Education Score}, 
				symbolic x coords={
					2019,
					2020,
					2021,
					2022,
				},  
				xtick=data,  
				nodes near coords,  
				nodes near coords align={vertical},  
				width=0.5\textwidth, 
				height=5cm, 
				]  
				\addplot [fill=yellow] coordinates {
					(2019,6)
					(2020,7)
					(2021,6.25)
					(2022,8.25)	
				};
				\addplot [fill=green] coordinates {
					(2019,7.37)
					(2020,8.14)
					(2021,8.37)
					(2022,8.03)			
				};
				\addplot [fill=blue] coordinates {
					(2019,7.75)
					(2020,8.75)
					(2021,9.25)
					(2022,8.5)
				};
				\legend{ChatGPT, AVS, MVS} 
			\end{axis}  
		\end{tikzpicture} 
	\end{center}
	\caption{Comparison in civic education.}
	\label{fig:social_civc}
\end{figure}

This section presents an investigation into ChatGPT performance in the civic education subject in comparison to Vietnamese students. The scores of ChatGPT and Vietnamese students from 2019-2022 are depicted in Figure~\ref{fig:social_civc}. Our results show that ChatGPT scores in the civic education subject ranged from 6.0 to 8.25, but consistently achieved lower results compared to the majority of Vietnamese students. Although in 2022, the gap between ChatGPT scores and Vietnamese students' scores was less significant than in previous years, ChatGPT still scored lower than most Vietnamese students.

One possible explanation for ChatGPT lower performance is its lack of analytical skills and higher-order thinking required for the problem-solving section of the civic education exam, which is a model of the law and how to apply it. While ChatGPT answered theoretical questions related to the law relatively well, it struggles with this type of question.

Our findings suggest that while ChatGPT has demonstrated impressive language-related abilities in various tasks, its performance in subject-specific tests still has room for improvement. Further studies are needed to identify the factors that contribute to ChatGPT performance in subject-specific tests and to develop strategies for enhancing its performance in these contexts. These findings are essential as it highlights the importance of developing ChatGPT analytical and problem-solving skills in the context of subject-specific tests. It also underscores the need for future developments in AI language models to improve their performance in tasks that require higher-order thinking skills.

\subsection{ChatGPT passes the Vietnamese National High School Graduation Examination}

Scoring formula of the Ministry of Education and Training to consider graduation:

\begin{equation} \label{eq1}
	\begin{split}
		\mathrm{G A S} & =\frac{\left(\frac{ \mathrm{T F S}+\mathrm{T I P}}{4}\right) \times 7+(\mathrm{A V 12}) \times 3}{10}+\mathrm{P P} 
	\end{split}
\end{equation}
In which:
GAS is General Admission Score;
$\mathrm{TFS} = \frac{\mathrm{M+L+E+A C}}{4}$ where TFS, M, L, E and AC are Total Four Subjects, Mathematics, Literature, English, and Averaged Combination scores, respectively;
TIP is Total Incentive Points;  
AV12 is Grade point average for the whole year of grade 12; 
PP is Priority Points; 
AC is Average of Combination, which includes two subcategories:
Average of natural combination $\mathrm{AC_N}  = \frac{1}{3}(\mathrm{Phy+Che+Bio})$; 
Average of social combination $\mathrm{AC_S}  = \frac{1}{3}(\mathrm{His+Geo+Civ})$. 
In case TIP, PP are not taken into account, and suppose that $\mathrm{AV12} = \frac{1}{4}(\mathrm{M+L+E+AC})$, we obtain
\begin{equation} \label{eq2}
\mathrm{GAS}=\frac{1}{4}(\mathrm{M}+\mathrm{L}+\mathrm{E}+\mathrm{AC})
\end{equation}

To evaluate ChatGPT performance in terms of graduation, we utilized the scoring formula of the Ministry of Education and Training and computed the combined scores for natural and social classes. Table~\ref{tab:GAS} presented the Natural and Social combination scores for each year from 2019 to 2023. The results showed that ChatGPT performance in completing the examination was sufficient to pass for all years, with combined scores ranging from 6.35 in 2019 to 6.94 in 2020. 

\begingroup
\renewcommand{\arraystretch}{1.5} 

\begin{table}[ht!]
	\caption{Subject scores and combination scores for natural and social classes}
	\label{tab:GAS}
	\centering

		\begin{tabular}{c|c|c|c|c|c|c|c|}
			\cline{2-8}
			& \textbf{Math} & \textbf{Lit} & \textbf{Eng} & \textbf{$\mathrm{AC_N}$} & \textbf{$\mathrm{AC_S}$} & \textbf{$\mathrm{GAS_N}$} & \textbf{$\mathrm{GAS_S}$} \\ \hline
			\multicolumn{1}{|c|}{\textbf{2019}} & 5.20          & 7.50         & 7.60        & 5.33            & 5.08            & 6.41             & 6.35             \\ \hline
			\multicolumn{1}{|c|}{\textbf{2020}} & 6.60          & 6.88         & 8.60        & 5.50            & 5.67            & 6.89             & 6.94             \\ \hline
			\multicolumn{1}{|c|}{\textbf{2021}} & 6.00          & 7.50         & 7.60        & 5.83            & 6.42            & 6.73             & 6.88             \\ \hline
			\multicolumn{1}{|c|}{\textbf{2022}} & 6.20          & 5.63         & 8.00        & 5.67            & 6.83            & 6.37             & 6.66             \\ \hline
			\multicolumn{1}{|c|}{\textbf{2023}} & 5.40          & 6.48         & 7.80        & 5.50            & 7.42            & 6.30             & 6.77             \\ \hline
		\end{tabular}

\end{table}
\endgroup

\begin{figure}[h!]
	\begin{center}
				\begin{tikzpicture}  
			\begin{axis}  
				[  
				ybar, 
				bar width=15pt,
				enlargelimits=0.15, 
				legend style={at={(0.5,-0.175)}, 	
					anchor=north,legend columns=-1},       
				ylabel={GAS}, 
				symbolic x coords={
					2019,
					2020,
					2021,
					2022,
					2023,
				},  
				xtick=data,  
				nodes near coords,  
				nodes near coords align={vertical},  
				width=\textwidth, 
				height=5cm, 
				width=9cm,
				]  

				\addplot [fill=red] coordinates {
					(2019,6.41)
					(2020,6.89)
					(2021,6.73)
					(2022,6.37)
					(2023,6.30)						
				};
	
				\addplot [fill=blue] coordinates {
					(2019,6.35)
					(2020,6.94)
					(2021,6.88)
					(2022,6.66)
					(2023,6.77)
				};
				\legend{$\mathrm{GAS_N}$, $\mathrm{GAS_S}$} 
			\end{axis}  
		\end{tikzpicture}
	\end{center}
	\caption{Graph of combination scores for natural and social classes from 2019 to 2023.}
	\label{fig:GAS}
\end{figure}
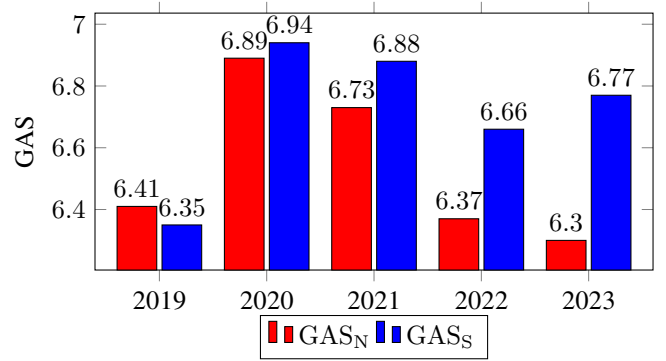
	
The graph presented in Figure~\ref{fig:GAS} displays the natural and social combination scores achieved by ChatGPT from 2019 to 2023. Our findings show that ChatGPT consistently performs well in both natural and social combinations, with no significant difference between the scores. This observation is noteworthy, as the VNHSGE exam includes a considerable number of theoretical questions that emphasize knowledge and comprehension. ChatGPT exceptional accuracy in answering these types of questions implies that the model holds considerable promise for passing the VNHSGE exam in the future. The model's consistent and stable performance over the years in both natural and social combinations highlight its versatility and training capabilities in various subjects. Moreover, the consistent performance of ChatGPT in both natural and social combinations may have significant implications for students preparing for the VNHSGE exam. ChatGPT can serve as an effective tool for students to enhance their knowledge and comprehension levels in both fields. While ChatGPT performance in subject-specific tests requires further improvements, its potential to answer general knowledge questions related to both fields with high accuracy could be beneficial to students in their exam preparations. Further studies are necessary to examine the factors contributing to ChatGPT consistent performance in both natural and social combinations. Overall, the study demonstrates that ChatGPT has shown significant promise as a valuable tool for students preparing for the VNHSGE exam.
	
\section{Discussion}

The findings of this study demonstrate that ChatGPT has great potential for use in various educational settings. The ability of ChatGPT to achieve an average score of 6-7 on the VNHSGE exam is an encouraging indication of its capabilities. Our research reveals that ChatGPT performed exceedingly well in mathematics, English, physics, chemistry, biology, history, geography, civic education, and literature, achieving scores similar to those attained by high school students. These results highlight the potential of ChatGPT to provide effective support for learners in different fields.

However, it is crucial to consider the types of examination questions and the extent of ChatGPT training in answering those specific types of questions. The VNHSGE exam evaluates students' abilities to apply critical thinking, analytical, and problem-solving skills across a diverse range of subjects, some of which require memorization of facts and formulas. While ChatGPT shows promise in its capacity to manage a range of subjects, further research is needed to evaluate its performance on more complex exam questions and its potential to support learners in various contexts.

The results of this study offer several implications for educators and researchers. First, ChatGPT may be utilized as an effective tool to support learners in various subjects. This AI model may serve as a reliable source of information, answering questions, and providing feedback. Second, educators may use ChatGPT as a complement to traditional teaching methods, providing students with personalized learning experiences. Third, researchers may employ ChatGPT to explore the nature of machine learning in education and develop new approaches for AI-based education.

Overall, ChatGPT potential for use in education is evident. With the ability to pass the VNHSGE exam and achieve comparable scores to those of high school students across a range of subjects, ChatGPT may serve as an essential tool for learners, educators, and researchers. However, further research is needed to evaluate the model's performance in more complex exam questions and its potential to support learners in diverse contexts.

\section{Conclusion}

In conclusion, this research article presents the use of ChatGPT for completing the VNHSGE exam, and the results showed that ChatGPT was able to pass the exam. This achievement demonstrates the potential of ChatGPT to automate the exam-taking process and provide a new approach for students to prepare for exams. However, further research is needed to explore the effectiveness of ChatGPT in different educational settings and to identify potential challenges and limitations.

Despite the need for further research, the possibilities for ChatGPT to transform the education landscape are promising. The application of AI in education is a field that is rapidly growing. With continued advancements in AI technology, we can expect to see more opportunities for ChatGPT and similar tools to be used in various educational contexts.

In summary, the successful completion of the VNHSGE exam by ChatGPT is a significant milestone in the development of AI in education. It highlights the potential of AI tools to provide new approaches for learning and assessment, and the need for further research to fully explore their capabilities and limitations.

\bibliographystyle{IEEEtran}
\bibliography{paper_7_9}

\end{document}